%% file: paper4_EIM.tex
\documentclass[onefignum,onetabnum]{siamart171218}
\input{ex_shared}
\usepackage{tikz}
\usepackage{blindtext}
\usepackage{lipsum}
\usepackage{multicol}
\usepackage{caption}
\usetikzlibrary{shapes, arrows}
\tikzstyle{Start} = [rectangle, minimum width=3cm, draw, text centered, minimum height=2em] 
\tikzstyle{terminator} = [rectangle, minimum width=3cm, draw, text centered,text width=5cm, minimum height=2em] 
\tikzstyle{process3} = [rectangle, minimum width=3cm, draw, text centered,text width=4cm, minimum height=2em] 
\tikzstyle{process4} = [rectangle, minimum width=3cm, draw, text centered,text width=3cm, minimum height=2em] 
\tikzstyle{connector} = [draw, -latex']
\tikzstyle{arrow} = [thick,->,>=stealth]
\usepackage[textsize=small]{todonotes}
\ifpdf
\hypersetup{
  pdftitle={An Example Article},
  pdfauthor={D. Doe, P. T. Frank, and J. E. Smith}
}
\fi

\begin{document}

\maketitle

\begin{abstract}
Empirical interpolation method (EIM) is a well-known technique to efficiently 
approximate parameterized functions. This paper proposes to use 
EIM algorithm to efficiently reduce the dimension of the training data within 
supervised machine learning. This is termed as DNN-EIM. Applications in data 
science (e.g., MNIST) and parameterized (and time-dependent) partial differential 
equations (PDEs) are considered. The proposed DNNs in case of classification are 
trained in parallel for each class. This approach is sequential, i.e., new classes can be added without having to retrain the network. In case of PDEs, a DNN is designed corresponding to each EIM point. Again, these networks can be trained in parallel, for each EIM point. In all cases, the parallel networks require fewer 
than ten times the number of training weights. Significant gains are observed in 
terms of training times, without sacrificing accuracy.
\end{abstract}


\section{Introduction}

Since its inception in \cite{MBarrault_YMaday_NCNguyen_ATPatera2004a}, see also \cite{YMaday_2009a,chaturantabut2010nonlinear}, the Empirical Interpolation Method 
(EIM) has been widely used as an interpolation technique to efficiently approximate 
parameterized functions. It has led to the design of new quadrature rules \cite{HAntil_SField_RHNochetto_MTiglio_2013} and applications in gravitational waves \cite{HAntil_DChen_SField_2018a}. Building on the success story of EIM, the goal 
of this paper is to use the EIM interpolation properties to reduce the dimension
of the training datasets in supervised machine learning. In particular, the 
focus will be on Deep Neural Networks (DNNs). 

Other reduced order modeling techniques, such as principal component analysis (PCA) or proper orthogonal decomposition 
(POD) and autoencoders, have been used to reduce the DNN training costs \cite{maulik2021reduced,salvador2021non,hesthaven2018non,franco2023deep,san2019artificial}. However, the projection based techniques such as POD or reduced basis method (RBM) require function evaluations in full space. In case of image classification problem, one needs to store the entire image to arrive at the input training data. Consider for instance, a vector valued parameterized function $\mathbf{f}(\eta) \in \mathbb{R}^n$ where $\eta$ is the parameter. Let the columns of $\mathbf{V} \in \mathbb{R}^{n \times m}$ contain the bases. Then the projection of $\mathbf{f}(\eta)$ onto the reduced space means computing $\mathbf{V}^T \mathbf{f}(\eta)$. This computation still requires you to evaluate $\mathbf{f}(\eta)$ for all $n$-components of $\mathbf{f}$. This can be expensive for a certain class of applications of parameterized functions \cite{HAntil_SField_RHNochetto_MTiglio_2013} and, also for image classification
problems where one cannot store the entire image dataset.

The proposed EIM based interpolation approach can be thought as an alternative 
to the projection based techniques, to reduce the dimensionality of data in 
DNN training, with comparable accuracy. Similarly to the existing approaches, 
it leads to reduced training times and prediction costs. Additional savings 
are observed because of the parallel nature of the proposed networks. Indeed, 
in case of classification problems, we consider one network per class. In
case of PDEs, we consider one network per EIM point. The proposed networks
need much smaller number of training weights. In combination with the 
parallel nature of the networks, significant time and memory savings are
observed, without sacrificing accuracy. Furthermore, the approach for 
classification problems is sequential in nature, i.e., more classes can
be added without having to retrain the network. 

The efficacy of the proposed approach is illustrated by multiple examples, 
including standard MNIST dataset, 1D (space-time) Kuramoto-Sivashinksky 
PDE and 2D parameterized advection dominated diffusion equation. 

\smallskip
\noindent
{\bf Outline:} Section \ref{sec:NN} contains the definitions which are necessary 
to introduce our algorithm. This include, basic workings of EIM and DNNs. 
Section \ref{sec:method} details our main DNN-EIM algorithm and all of its 
components. Section~\ref{sec:numericalresults} discusses the implementation
details, followed by multiple examples including MNIST dataset, the 
1D (space-time) Kuramoto--Sivashinsky equation, and 2D parameterized 
advection diffusion equation.

\section{Preliminaries}
\label{sec:NN}
The purpose of this section is to describe the empirical interpolation method (EIM) in section \ref{sec:deim} and Residual Neural Networks (ResNets) in section \ref{sec:resnets}. The material in this section is well known.

\subsection{Empirical Interpolation Method}\label{sec:deim}

We begin by describing the empirical interpolation method (EIM) for vector valued 
parameterized functions $\mathbf{u}:\mathcal{D} \rightarrow \mathbb{R}^n$, with 
$\mathcal{D}\subset\mathbb{R}^d$ being a compact set \cite{MBarrault_YMaday_NCNguyen_ATPatera2004a,chaturantabut2010nonlinear}. 
The goal is to approximate $\mathbf{u}$ in a lower dimensional subspace spanned 
by a linearly independent set $\{\mathbf{v}_1,\dots,\mathbf{v}_m\}\subset\mathbb{R}^n$ 
with $m\ll n$. In other words, for $ \eta \in \mathcal{D}$ 
\begin{equation*}
    \mathbf{u}(\eta)\approx \mathbf{V} \mathbf{c}(\eta),
\end{equation*}
where $\mathbf{V}=[\mathbf{v}_1,\dots,\mathbf{v}_m]\in\mathbb{R}^{n\times m}$ and $\mathbf{c}(\eta)\in\mathbb{R}^m$ is the corresponding coefficient vector. The linear system $\mathbf{u} = \mathbf{V}\mathbf{c}$ is overdetermined unless we select $m$ rows designated by the indices $\{p_1,\dots,p_m\}$. Let $\mathbf{e}_{p_i}$ be the $\mathbb{R}^n$ standard basis vector with a 1 in the $p_i$-th component. Let $\mathbf{P}\in\mathbb{R}^{n\times m}$ be the matrix with columns $\mathbf{e}_{p_i}$. If $\mathbf{P}^T\mathbf{V}$ is nonsingular then 
\begin{equation*}
    \mathbf{P}^T\mathbf{u}=\mathbf{P}^T\mathbf{V}\mathbf{c} \, ,
\end{equation*}
will have a unique solution $\mathbf{c}$. Then $\mathbf{u}(\eta)$ is approximated as
\begin{equation*}
    \mathbf{u}(\eta) \approx \hat{\mathbf{u}} (\eta)  :=\mathbf{V}(\mathbf{P}^T\mathbf{V})^{-1}\mathbf{P}^T\mathbf{u}(\eta).
\end{equation*}
For approximation error estimates between $\mathbf{u}$ and
$\hat{\mathbf{u}}$, we refer to \cite{chaturantabut2010nonlinear}. 
Notice that, this approximation requires a projection basis 
$\{\mathbf{v}_1,\dots,\mathbf{v}_m\}$ and interpolation indices 
$\{p_1,\dots,p_m\}$. The projection basis is determined by applying 
a reduced order modeling technique such as proper orthogonal
decomposition (POD) or reduced basis method (RBM) 
\cite{JSHesthaven_GRozza_BStamm_2016a}, or matrix sketching 
\cite{MR3732946}. 
The interpolation indices are determined 
from the EIM algorithm and we will refer to them as the EIM points. 
To summarize and for completeness, we restate Definition 3.1 
from \cite{chaturantabut2010nonlinear}.

\begin{definition}[EIM Approximation]\label{def:DEIM}
Let $\mathbf{u}:\mathcal{D}\rightarrow \mathbb{R}^n$ be a nonlinear vector-valued function with $\mathcal{D}\subset\mathbb{R}^d$ for some positive integer $d$. Let $\{ \mathbf{v}_\ell\}_{\ell=1}^m\subset\mathbb{R}^n$ be a linearly independent set for $m\in\{1,\dots,n\}.$ For $\eta\in\mathcal{D}$, the EIM approximation of order m for $\mathbf{u}(\eta)$ in the space spanned by $\{\mathbf{v}_\ell\}_{\ell=1}^m$ is given by
\begin{equation}\label{eq:DEIMapprox}
    \hat{\mathbf{u}}(\eta):= \mathbf{V}(\mathbf{P}^T\mathbf{V})^{-1}\mathbf{P}^T\mathbf{u}(\eta),
\end{equation}
where $\mathbf{V}=[\mathbf{v}_1,\dots,\mathbf{v}_m]\in\mathbb{R}^{n\times m}$ and $\mathbf{P} = [\mathbf{e}_{p_1},\dots,\mathbf{e}_{p_m}]\in\mathbb{R}^{n\times m}$, with $\{p_1,\dots,p_m\}$ being the output from the EIM algorithm with the input basis $\{\mathbf{v}_i\}_{i=1}^m$.    
\end{definition}

\subsection{Residual Neural Networks}\label{sec:resnets}
The neural networks of choice in this work are ResNets 
\cite{he2016deep,LRuthotto_EHaber_2019a}. 
The components involved are the input vector $\xi \in\mathbb{R}^d$, 
inner layer vectors $f_\ell\in\mathbb{R}^{n_\ell}$, and the output 
vector $f_L\in\mathbb{R}^{d^*}$. Here, $d,d^*$, and $n_\ell$ are 
positive, independent integers. Additional components include 
the weight matrices $W_\ell\in\mathbb{R}^{n_\ell\times n_{\ell+1}}$ 
and the bias vectors $b_\ell\in\mathbb{R}^{n_{\ell +1}}$ where the 
$\ell$-th layer has $n_\ell$ neurons. A typical ResNet has the 
following structure:
\begin{align}\label{ResNet}
\begin{aligned}
    f_1 &:= \sigma(W_0 \xi +b_0),\\ 
    f_{\ell+1} &:= f_\ell + \tau\sigma(W_\ell f_\ell + b_\ell),\quad \ell=1,...,L-2,\\ 
    f_L &:= W_{L-1}y_{L-1} .
\end{aligned}    
\end{align}
The scalar $\tau>0$ and the activation function $\sigma$ are user defined. For the purpose of this work, we have chosen a smooth quadratic approximation of the ReLU function,
\begin{equation*}\label{sigma}
 \sigma(x) =
    \begin{cases}
     \max\{0,x\} & |x|>\epsilon,\\
      \frac{1}{4\epsilon}x^2+\frac{1}{2}x+\frac{\epsilon}{4} & |x|\le\epsilon.\\
    \end{cases}     
\end{equation*}

First considered in \cite{antil2021deep}, we introduce the following optimization problem which represents the  ResNet \eqref{ResNet} training optimization problem. The additional components of the optimization problem are the loss function $J$ and the training (input-output) data $\{( \xi^i,y^i)\}_{i=1}^{N_s}$. The ResNet output is calculated via a map $\mathcal{F}$. The learning problem is given by
\begin{subequations}
\begin{align}\label{NN}
    \min_{\{W_\ell\}_{\ell=0}^{L-1},\{b_\ell\}_{\ell=0}^{L-2}} J(\{(f_L^i,y^i)\}_i,\{W_\ell\}_\ell,\{b_\ell\}_\ell), \\
    \text{subject to }f_L^i = \mathcal{F}(\xi^i;(\{W_\ell\},\{b_\ell\})),\quad i=1,...,N_s, \label{DNN} \\
    b_\ell^j\le b_\ell^{j+1},\quad j=1,...,n_{\ell+1}-1,\quad \ell=0,...,L-2. \label{biasordering} 
\end{align}
\end{subequations}
In our numerical experiments the loss function $J$ in \eqref{NN} will be quadratic 
\begin{align}\label{lossf}
J:=\frac{1}{2 N_s}\sum_{i=1}^{N_s}\|f_L^i-y^i\|_2^2 + \frac{\lambda}{2}\sum_{\ell=0}^{L-1} \left(\|W_\ell\|_1 + \|b_\ell\|_1 + \|W_\ell\|_2^2 + \|b_\ell\|_2^2 \right) ,
\end{align}
where $\lambda\ge0$ is the regularization parameter. Note that there is no $b_{L-1}$ term here due to the absence of the bias at final layer in the ResNet \eqref{ResNet}.
The second summation regularizes the weights and biases using $\ell_1$ and $\ell_2$ norms. Following \cite{antil2021deep}, and motivated by Moreau-Yosida regularization, the bias ordering \eqref{biasordering} is implemented as an additional penalty term in $J$
\begin{equation}\label{biasorderingpenalty}
    J_{\gamma} := J + \frac{\gamma}{2}\sum_{\ell=0}^{L-2}\sum_{j=1}^{n_{\ell+1}-1}\|\min\{b_{\ell}^{j+1}-b_{\ell}^j,0\}\|_2^2 .
\end{equation}
Here $\gamma$ is a penalization parameter and for convergence results as $\gamma \rightarrow \infty$, see \cite{antil2021deep}.

\section{Method}
\label{sec:method}

This section introduces the \emph{DNN-EIM} algorithm. 
In section \ref{sec:offlinealgorithm} we discuss the DNN-EIM algorithm which 
includes generating the EIM interpolation indices to reduce the 
dimension of the training data and subsequently training the 
neural network surrogate. This is an offline step. 

\subsection{DNN-EIM Algorithm}
\label{sec:offlinealgorithm}
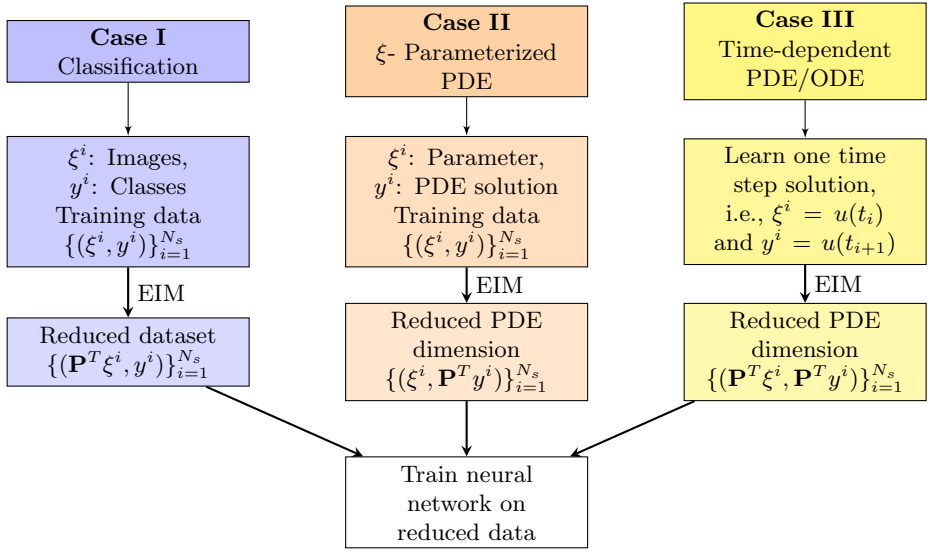
\begin{figure}
\small
\begin{center}
\begin{tikzpicture}
\node [process4, fill=blue!25] at (-7,-1.5) (Case I) { \textbf{Case I} \\ Classification};
\node [process4, fill=blue!20] at (-7,-3.5) (Goal Case I) {$\xi^i$: Images, $y^i$: Classes\\Training data $\{(\xi^i,y^i)\}_{i=1}^{N_s}$};
\path [connector] (Case I) -- (Goal Case I);
\node [process4, fill=blue!15] at (-7,-5.5) (DEIM I) {Reduced dataset \\$\{(\mathbf{P}^T\xi^i, y^i)\}_{i=1}^{N_s}$};
\draw [arrow] (Goal Case I) -- node[anchor=west]  {EIM} (DEIM I);
\node [process4, fill=orange!35] at (-2.5,-1.5) (Case II) { \textbf{Case II} \\ $\xi$- Parameterized\\ PDE};
\node [process4, fill=orange!25] at (-2.5,-3.5) (Goal Case II) {$\xi^i$: Parameter, $y^i$: PDE solution\\Training data $\{(\xi^i,y^i)\}_{i=1}^{N_s}$};
\path [connector] (Case II) -- (Goal Case II);
\node [process4, fill=orange!20] at (-2.5,-5.5) (DEIM II) {Reduced PDE dimension \\$\{(\xi^i,\mathbf{P}^T y^i)\}_{i=1}^{N_s}$};
\draw [arrow] (Goal Case II) -- node[anchor=west]  {EIM} (DEIM II);
\node [process4]
at (-2.5,-7.5) (NN II) {Train neural\\ network on\\ reduced data};
 \draw [arrow] (DEIM II) --  (NN II);
\node [process4, fill=yellow!60] at (2,-1.5) (Case III) {\textbf{Case III}\\ Time-dependent PDE/ODE};
\node [process4, fill=yellow!50] at (2,-3.5) (Goal Case III) {Learn one time step solution,\\ i.e., $\xi^i = u(t_i)$ and $y^i=u(t_{i+1})$};
\path [connector] (Case III) -- (Goal Case III);
\node [process4, fill=yellow!40] at (2,-5.5) (DEIM III) {Reduced PDE dimension\\ $\{(\mathbf{P}^T \xi^i,\mathbf{P}^T y^i)\}_{i=1}^{N_s}$};
 \draw [arrow] (Goal Case III) -- node[anchor=west]  {EIM} (DEIM III);
 \draw [arrow] (DEIM III) --  (NN II);
 \draw [arrow] (DEIM I) --  (NN II);
\end{tikzpicture}
\end{center}
\caption{Visualization of Algorithm \ref{alg:offline} for three cases. First column corresponds to generic supervised learning (including 
classification). Second column considers a parameterized PDE and the third 
column consists of a time dependent PDE. Notice that the right column can be 
directly extended to include additional parameters.}
\label{fig: tikz_Cases}
\end{figure}

The Algorithm~\ref{alg:offline} consists of five steps.
\begin{algorithm}
\caption{DNN-EIM Algorithm (Offline Stage)}\label{alg:offline}
\begin{algorithmic}[1]
\STATE Generate training and testing data $\{ (\xi^i, y^i) \}_{i=1}^{N_s}$.
\STATE Determine whether the input $\xi^i$, output $y^i$ or both will be reduced.
\STATE Determine a projection basis $\{\mathbf{v}_1,\dots,\mathbf{v}_m\}\subset\mathbb{R}^n$.
\STATE Apply EIM algorithm \cite{chaturantabut2010nonlinear} to determine interpolation indices $\{p_1,\dots,p_m\}$. 
\STATE Train neural network \eqref{NN} on the reduced training data. 
\end{algorithmic}
\end{algorithm}
A visualization of the algorithm is depicted in 
Figure~\ref{fig: tikz_Cases} for three generic scenarios 
which will be described next.

\textit{Step 1: Generate data.} 
This algorithm is applicable to a variety of training datasets.
Our focus is on three generic scenarios. Let $\xi^i \in \mathbb{R}^d$ 
and $y^i \in \mathbb{R}^n$ denote the input-output data 
$\{ (\xi^i, y^i) \}_{i=1}^{N_s}$. 
In the first case, we consider a generic dataset from the supervised machine learning. 
For instance, classification problems where $\xi^i$ and $y^i$, 
respectively represents the input images and output classes.
In the the second case, we consider a $\xi$-parameterized PDE, where 
the DNN aims to learn parameter 
$\xi^i$ to PDE solution $y^i$ map \cite{HAntil_HCElman_AOnwunta_DVerma_2021a}. 
Finally, we consider a time-dependent PDE/ODE with solution $u$.   
Here we aim to learn one time step solution. 
Let the solution at the previous time-step be $\xi^i = u(t_i)$, 
then we aim to learn $y^i=u(t_{i+1})$ \cite{brown2021novel}.  
Notice that in the third case the algorithm can be 
directly extended to incorporate additional parameters.

\textit{Step 2: Determine whether the input $\xi^i$, output $y^i$ or both 
will be reduced.} This step is problem dependent and here we describe it 
for the three specific cases shown in Figure \ref{fig: tikz_Cases}. 
\begin{enumerate}[$\bullet$]
\item \textbf{Classification:} In a classification problem, we reduce the dimension 
    of input $\xi^i$. 
 \item \textbf{Learning parameter to PDE solution map:} 
    In this case, we reduce the output $y^i$ as it corresponds to a PDE solution.
 \item \textbf{Learning one timestep PDE/ODE solution:} In this case, we reduce the input $\xi^i$ and the output $y^i$ as both corresponds to a PDE solution at time $t_i$ and $t_{i+1}$ respectively.
\end{enumerate}

\textit{Step 3: Generate projection basis} either using POD, RBM 
or matrix sketching. Let the resulting bases be given by  
$\{\mathbf{v}_1,\dots,\mathbf{v}_m\}\subset\mathbb{R}^n$. 

\textit{Step 4: Apply EIM algorithm to determine the interpolation indices \cite[Algorithm 1]{chaturantabut2010nonlinear},} namely 
\begin{equation*}
    \{p_1,\dots,p_m\} = EIM ( \{\mathbf{v}_1,\dots,\mathbf{v}_m\}).
\end{equation*}
The output of the EIM algorithm are the interpolation indices 
$\{p_1,\dots,p_m\}$ or EIM points. We set $\mathbf{P} = [\mathbf{e}_{p_1},\dots,\mathbf{e}_{p_m}]\in\mathbb{R}^{n\times m}$
and next describe the DNN training datasets for all these cases:
\begin{enumerate}[$\bullet$]
    \item \textbf{Classification}: The training data is $\{(\mathbf{P}^T\xi^i, y^i) \}_{i=1}^{N_s}$. Note that 
    the output variable $y^i$ represents the category/class in classification datasets.

    \item {\bf Learning parameter to PDE solution map:}  We set the training data to be $\{ (\xi^i, \mathbf{P}^Ty^i) \}_{i=1}^{N_s}$. Recall that the action of $\mathbf{P}^T$ on $y^i$ will pick the rows of $y^i$ corresponding to the EIM points.
    
    \item {\bf Learning one timestep PDE/ODE solution:} Recall that in this setting $\xi^i = u(t_i)$ and $y^i = u(t_{i+1})$, where $u$ solves the PDE/ODE. The training data here is given by 
    $\{ (\mathbf{P}^T\xi^i, \mathbf{P}^Ty^i) \}_{i=1}^{N_s}$.    
\end{enumerate}

The neural network operates on these EIM points as shown in Figure \ref{fig: tikz_Cases}.\\

\textit{Step 5: Train neural network.} 
Next, we parallelize our neural network by training multiple smaller ResNets. For classification, we will setup one network per class.
For the PDE examples, the `localness' of the PDE and the EIM 
interpolation is leveraged to reduce the training time and the number 
of parameters to learn. We describe our approach in three cases next.
\begin{enumerate}[$\bullet$]
\item \textbf{Classification:} We train one DNN per class.
    This allows the flexibility to add additional classes without relearning 
    the previous classes.
     \item \textbf{Learning parameter to PDE solution map:}  
    We train one DNN per EIM point, namely, the training data corresponding to the $j$-th EIM point is $\{ (\xi^i, \mathbf{e}_j^Ty^i) \}_{i=1}^{N_s}$. This leads to a highly efficient and embarrassingly parallel training algorithm.

\begin{multicols}{2}[\columnsep2em] 
\item {\bf Learning one timestep PDE/ODE solution:} 
Let $\{p^{1}_{j},\dots,p^{n_j}_{j}\}$ be a subset of EIM points that contains $p_j$ and the $n_j-1$ closest EIM points to $p_j$ in terms of Euclidean distance as shown in Figure \ref{fig: EIM_nbd} and set $\mathbf{P}_j = [\mathbf{e}_{p^{1}_{j}},\dots,\mathbf{e}_{p^{n_j}_{j}}]\in\mathbb{R}^{n_j\times m}$. Then the training data for the  $j$-th EIM point with input size $n_j$ is

\columnbreak

\includegraphics[trim=10 10 15 10,clip, width=0.8\linewidth]{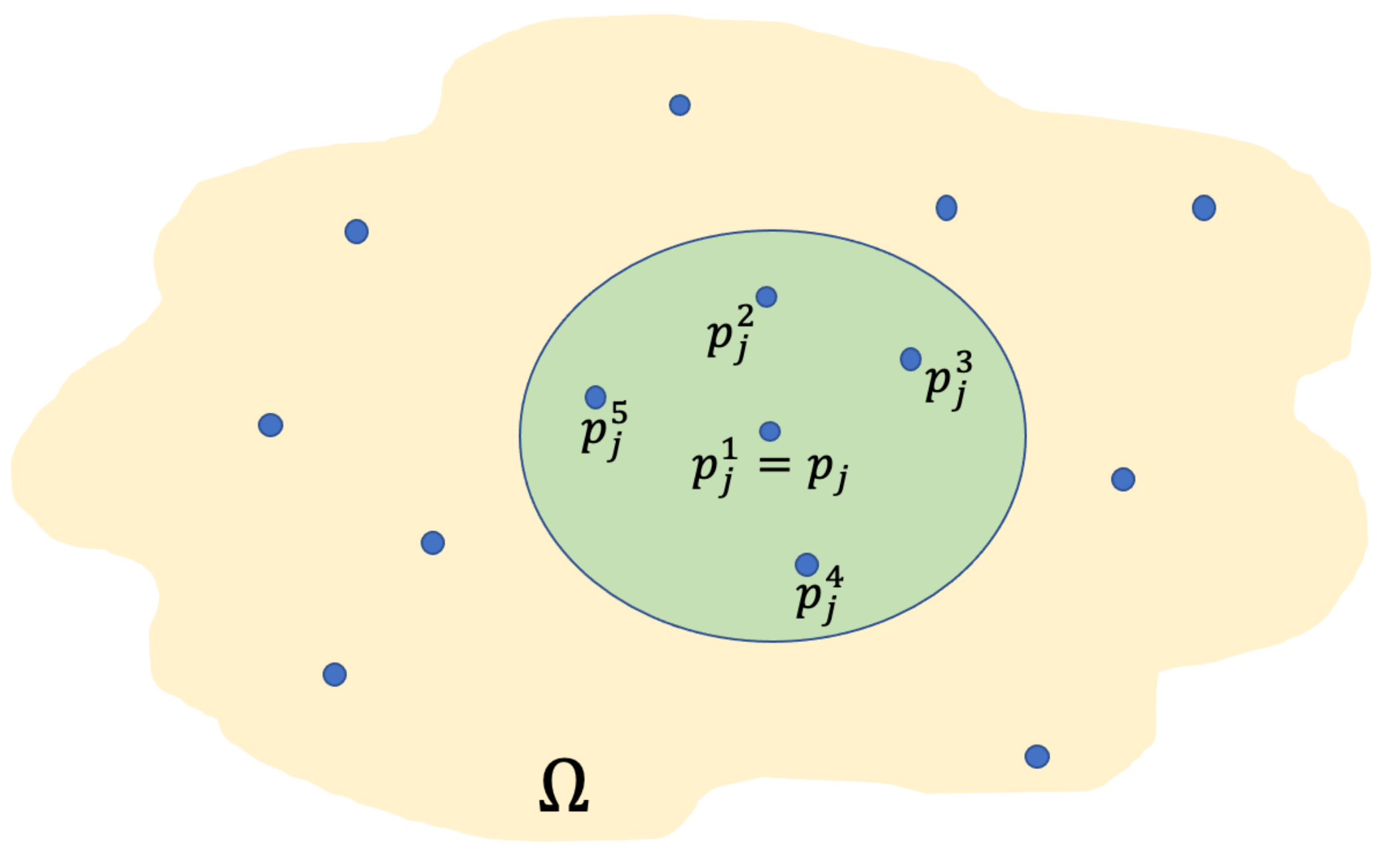}
\captionof{figure}{Neighbourhood of EIM points}\label{fig: EIM_nbd}

\end{multicols}

\end{enumerate}
   $\{ (\mathbf{P}_j^T\xi^i, \mathbf{e}_j^Ty^i) \}_{i=1}^{N_s}$.
Next, we demonstrate the application of the above algorithm to the MNIST 
data set. This will serve as a guide to help in understanding the more 
demanding PDE examples.

\subsection{MNIST Dataset}\label{sec: MNIST}

We explain our approach with the help of a specific MNIST dataset. However, this generic approach is applicable to any of the existing training datasets. The MNIST database consists of pairs of handwritten 
digit images and labels. The digits range from 0 to 9 and the dataset is comprised of 70,000 images. The algorithmic steps are:
\begin{enumerate}
    \item \textit{Generate data}: The MNIST dataset is available at \\ \textit{http://yann.lecun.com/exdb/mnist/}.
    \item \textit{Determine whether the input, output or both will be reduced}: This is a classification problem which implies the input to the neural network is the image and the output is the corresponding digit $0-9$. Therefore we will reduce the dimension of the input.
    \item \textit{Generate projection basis}: Next, we transform the $28\times28$ dimension images into $784$ dimension vectors. Then we gather them into a $784\times 70000$ dimension matrix. POD is applied to the matrix to generate a basis. Other approaches are equally applicable.
    \item \textit{Apply EIM algorithm to determine interpolation indices}: After the EIM algorithm is applied to the basis from step 3, the interpolation indices dictate which pixels are most important for classification purposes.
    \item \textit{Train neural network}: Finally, we train one ResNet per 
    class or digit. The input to the ResNet is the image at the 
    interpolation indices and the output of the ResNet is either 0 or 1 
    corresponding to a particular class. We refer to this system of 
    ResNets as DNN-EIM in Table \ref{tab:comparison}. 
\end{enumerate}
\medskip
We train a ResNet system as described above using Steps 1-5. We denote 
the system by 10 ResNets (corresponding to each class), to be trained in
parallel, as DNN-EIM and it uses 50 EIM points for the input. We compare 
DNN-EIM against POD and DNN-EIM (s). In POD, we are projecting the images 
onto a POD basis of dimension 50, and training a single ResNet on the projected data. By DNN-EIM (s), we are training a single ResNet on the EIM reduced dataset. For the POD and DNN-EIM (s) cases, we are using 50,000 samples for training, 10,000 for validation, and 10,000 for testing. 

For the DNN-EIM case, each individual ResNet is responsible for one digit. 
Then each of these ResNets are trained on 10,000 of the training samples 
and 2000 of the validation samples. These ResNets have 5000 training 
samples  which correspond to their assigned digit and 5000 training 
samples which correspond to other digits (chosen randomly from 50,000) to 
make up the total 10,000 training samples. The same split is applied to the 
2000 validation samples (1000 corresponds to the assigned digit and the 
remaining 1000 chosen randomly from 10,000). The testing is still done on the
entire test dataset. 

\begin{table}[h]
\centering
\resizebox{\textwidth}{!}
{
\begin{tabular}{|c|c|c|c|c|c|}
\hline
Methods &No. of ResNets & Layers & Width & No. of param. & Accuracy \\ [1.5ex]
\hline
\hline
POD  & 1 & 3 & 50 & 8,150 & $95.51 \%$\\
\hline
DNN-EIM (s) & 1 & 3 & 50 & 8,150 & $87.65 \%$ \\
\hline
DNN-EIM & 10 & 3 & 10 & 740 (per ResNet)  & $90.65 \%$  \\
\hline

\hline
\end{tabular}}
\caption{Comparison of POD, DNN-EIM (s) with single ResNet and DNN-EIM
with 10 parallel ReNets corresponding to each class. All three 
methods have a reduced space dimension of 50. However, DNN-EIM overall uses 
fewer parameters and much smaller number of parameters per network.
Recall that DNN-EIM is trained in parallel corresponding to each class.}
\label{tab:comparison}
\end{table}

We notice several benefits of using DNN-EIM over POD and 
DNN-EIM (s). 
Firstly, in case of POD, we need the entire image to be able to 
arrive at the input data via projection onto the POD basis. On 
the other hand, DNN-EIM only requires you to `evaluate' the image
at the EIM points. For the MNIST dataset, we consider 50 EIM points 
(image pixels). 
Secondly, DNN-EIM is more accurate than DNN-EIM (s).
Finally, DNN-EIM requires only 740 parameters per network which 
can be trained in parallel. This is more than 10 times smaller 
than the other two networks. 
The remaining sections of the paper, focus on the parameterized
and time-dependent PDEs.

\section{Numerical Results}
\label{sec:numericalresults}
Before discussing the actual examples, we provide some implementation
details.

We first focus on the implementation of Step 5 in 
Algorithm \ref{alg:offline}. 
We train ResNet~\eqref{ResNet} with the loss function and activation function 
specified in section \ref{sec:resnets}. The scalar $\tau$ from the ResNet 
formulation \eqref{ResNet} is set to $\tau=\frac{2}{L}$ where $L$ is the 
number of layers. The regularization parameter $\lambda$ from the loss 
function \eqref{lossf} is set to $\lambda=10^{-7}$. The bias ordering 
parameter $\gamma$ from the penalty term \eqref{biasorderingpenalty} is 
set to $\gamma=1000$. For the semilinear advection-diffusion-reaction 
equation we used $2,500$ training samples with $80\%$ used for training 
and $20\%$ used for validation. For the 1D Kuramoto-Sivashinsky equations, 
we used $38,000$ training samples with $80\%$ used for training and $20\%$ 
used for validation. The training samples for the 1D Kuramoto-Sivashinsky example are scaled such that the input and output components take values 
between $0$ and $1$. The patience is set at 400 which means training will 
continue for 400 additional iterations if the validation error increases. 
The full training optimization problem \eqref{NN}-\eqref{biasordering} is 
solved using BFGS and initialized with box initialization \cite{box}.

The time averaged absolute error is computed as the approximation of the integral 
\begin{equation}\label{TARE}
    \| w_{alg} -u^{ref} \|_{\rm error} 
        = \frac1T\left( \int_{0}^{T} \|w_{alg}(t)-u^{ref}(t)\|_{L^2(\Omega)}dt \right) ,
\end{equation}
where $w_{alg}$ corresponds to the underlying algorithm (DNN-EIM, for instance).

Next, we will apply the proposed algorithm to two examples: 
first to the semilinear advection-diffusion reaction equation 
in section \ref{sec: Adve_Diff}, and second to the 1D 
Kuramoto-Sivashinsky equation in section \ref{sec:KSE}.

\subsection{Semilinear Advection-Diffusion-Reaction PDE}
\label{sec: Adve_Diff}
\begin{figure}[h!]
    \centering
    \includegraphics[width=\textwidth]{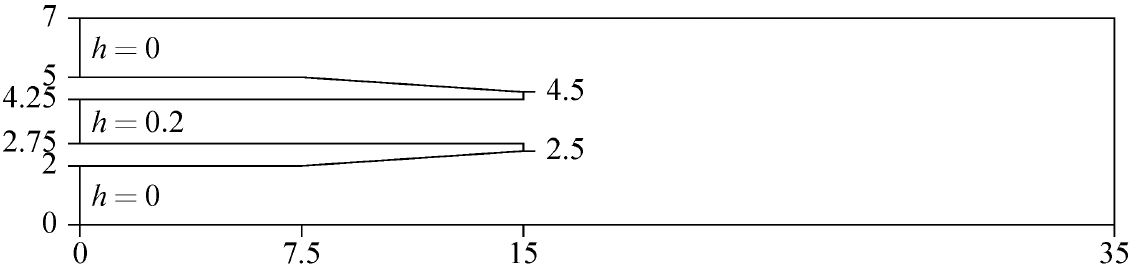}
    \caption{Physical domain $\Omega$ for the advection diffusion reaction
            example. See~\cite{antil2014application} for further details.}
    \label{f:domadv}
\end{figure}

The 2D semilinear advection diffusion reaction equation  with mixed boundary conditions is given by:
\begin{subequations}
\begin{align}\label{eq: advec diffusion}
    -\nabla \cdot (\nu \nabla u) + \beta \cdot \nabla u + f(u, \theta) &=0, \quad \text{in}~ \Omega,\\
    u&=h, \quad \text{on}~ \Gamma_D, \\
    \nabla u \cdot \mathbf{n} &=0, \quad \text{on}~ \Gamma_N.
\end{align}
\end{subequations}
This problem arises in a combustion process. Following 
\cite{antil2014application,galbally2010non}, we consider the 
following non-linearity
\begin{equation}
    f(y,\theta)=Ay(C-y)e^{-E/(D-y)},
\end{equation}
where $A,C,D,E\in\mathbb{R}$ are scalar parameters.
The domain $\Omega\subset\mathbb{R}^2$ is taken from \cite{antil2014application,becker2004numerical}
and is shown in Figure~\ref{f:domadv}
On $\Gamma_D$ and $\Gamma_N$, respectively we specify Dirichlet and Neumann boundary conditions. 
The diffusion coefficient is $\nu=5\cdot10^{-6}$ and the advection vector $\beta$ is 
constant with value $0.2$ in each direction. The nonlinear term parameters $C=0.2$ and $D=0.4$ 
are fixed. The remaining nonlinear parameters are grouped into $\theta=(\ln(A),E)$ and vary 
within $\Theta=[5.0,7.25]\times[0.05,0.15]\subset\mathbb{R}^2$. The Dirichlet boundary is 
given by $\Gamma_D=\{(0,x_2):x_2\in(0,2)\cup(2.75,4.25)\cup(5,7)\}$ on which we set $h = 0, 0.2~ \text{ and }~ 0$ respectively. We discretize 
the weak form of \eqref{eq: advec diffusion} using piecewise linear finite
elements with SUPG stabilization. 

The training samples for Step~1 of Algorithm \ref{alg:offline} are generated by selecting 50 equidistant parameters from both parameters $\Theta=[5.0,7.25]\times[0.05,0.15]\subset\mathbb{R}^2$ and solving the 
corresponding parameterized PDE. This results in 2500 solution snapshots.  
We compute a reference finite element solution by considering a parameter 
$\theta=(\ln(A),E)=(6.4,0.11)$ which is not part of training parameters. 

 \begin{figure}[!h]
\begin{center}
\includegraphics[width=.5\textwidth]{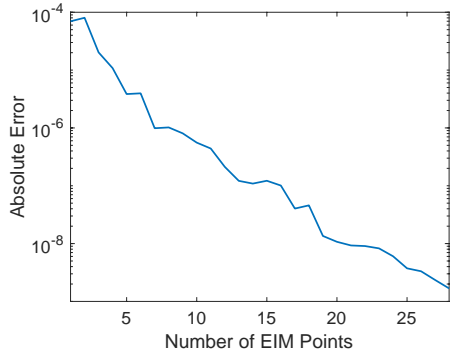}
\end{center}
\caption{EIM interpolation $L^2$-error for the semilinear advection-diffusion-reaction equation. The error is between the EIM approximation 
and a reference solution.}%
\label{fig:adv_diff_DEIM_err}%
\end{figure}

 \begin{figure}[!h]
\begin{center}
\includegraphics[trim=50 140 40 150,clip,width=\textwidth]{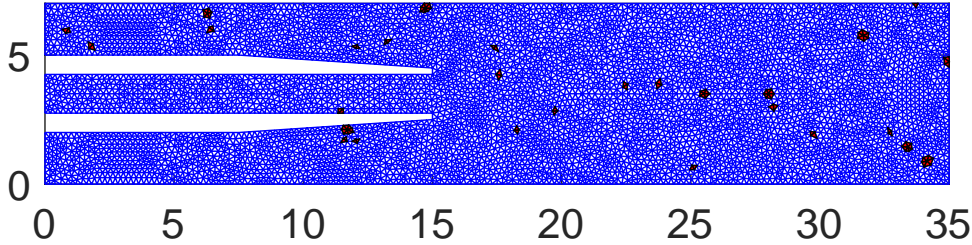}
\end{center}
\caption{EIM interpolation index locations indicated by red 
finite element triangles. The training output data that we 
tries to match in loss function corresponds to these marked 
triangles (cf.~\ref{fig: tikz_Cases}).}%
\label{fig:adv_diff_DEIM_loc}%
\end{figure}

\subsubsection{Neural Network Architecture}
\label{sec:nnarchadvdiff}
In this section, we will cover the structure of our ResNet system. For each EIM 
interpolation index $p_i$ generated by Step 3 of Algorithm \ref{alg:offline} 
we train one ResNet. The input to the ResNets are the values of the two parameters 
$\theta=(\ln(A),E)$. The output is the value of the solution at $p_i$, the $i$-th 
interpolation index generated by EIM in Step 4 of Algorithm \ref{alg:offline}.

The error plot pictured in Figure \ref{fig:adv_diff_DEIM_err} shows
the error between the EIM approximated solution and the reference solution 
both evaluated at $\theta=(\ln(A),E) = (6.4,0.11)$. This informs our choice 
of 28 EIM points as we see approximately $10^{-9}$ absolute interpolation 
error. The location of the 28 EIM points corresponding to the output data, to 
be matched with the DNN output, is pictured in Figure \ref{fig:adv_diff_DEIM_loc}. 

\subsubsection{Surrogate Quality}
\begin{figure}[h!]
\begin{center}
\includegraphics[trim=10 120 10 100,clip,width=.8\textwidth]{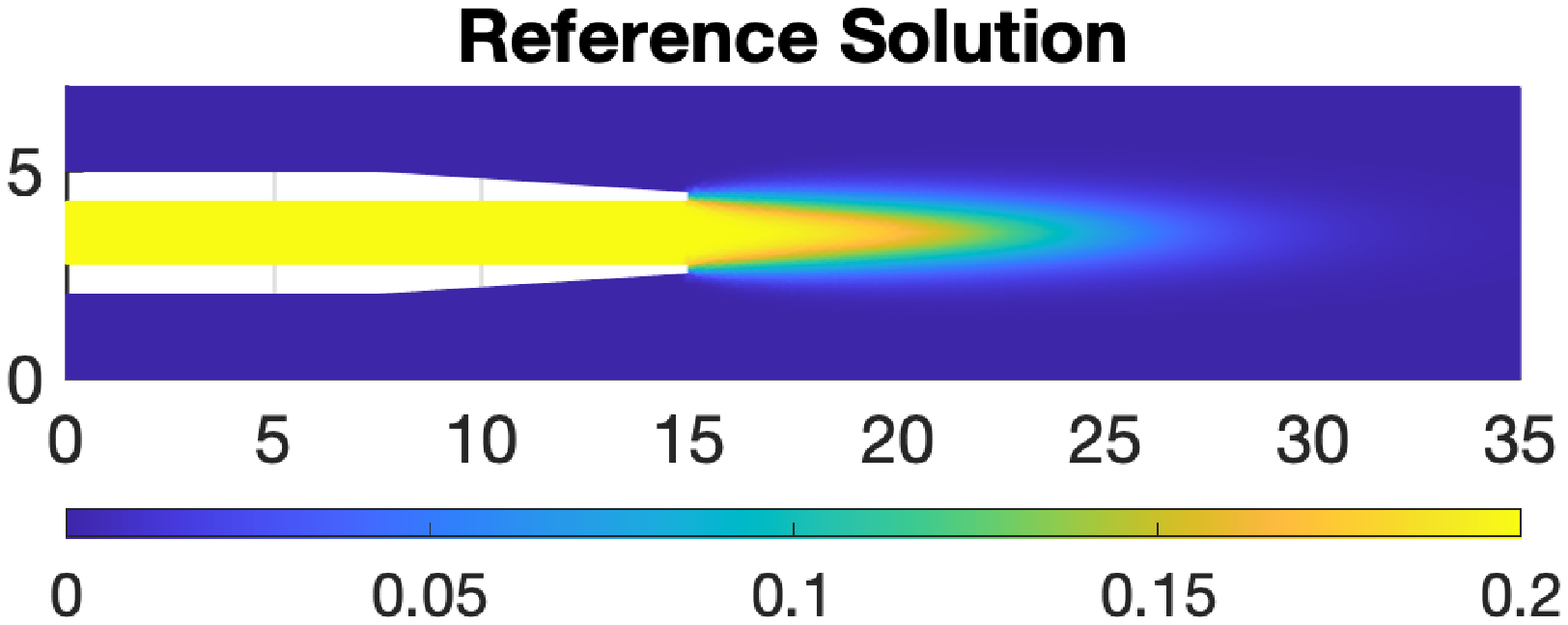}
\includegraphics[trim=10 120 10 90,clip,width=.8\textwidth]{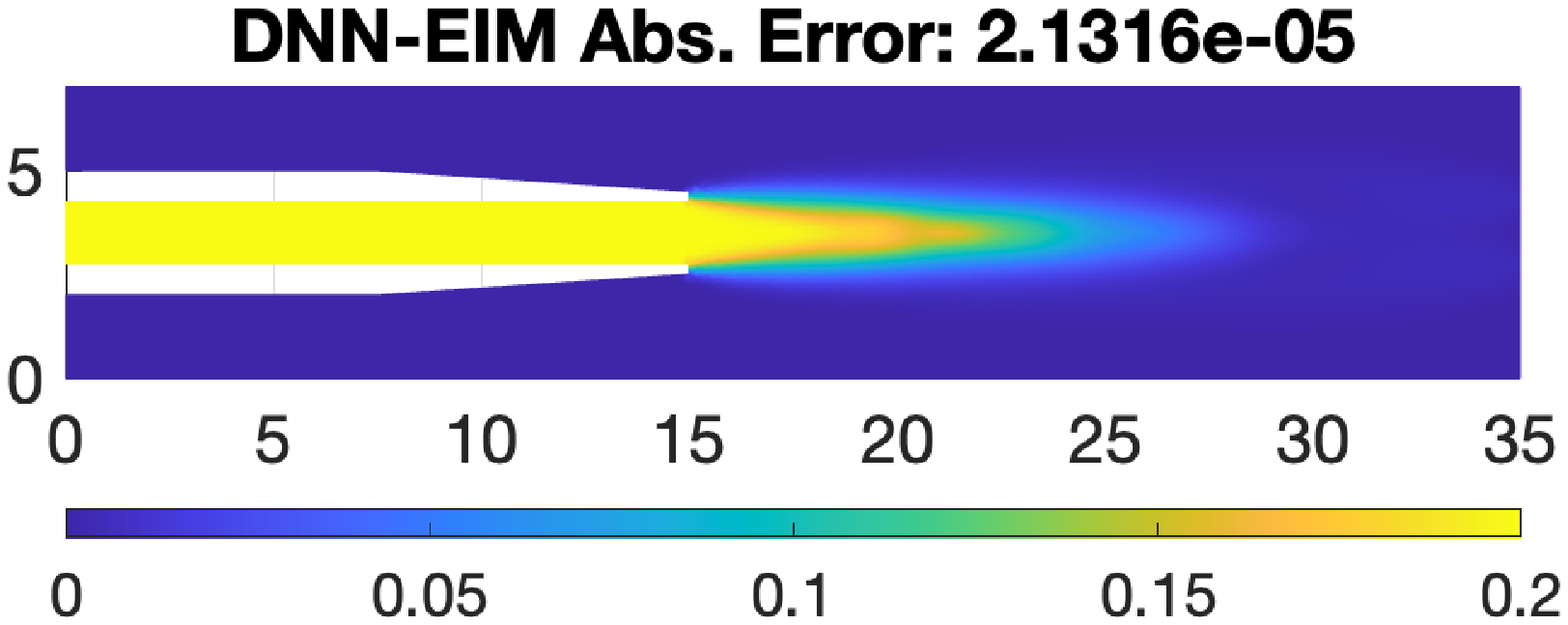}
\end{center}
\caption{Top: Reference (finite element) solution, Bottom: DNN-EIM solution for the semilinear advection-diffusion-reaction equation. The DNN-EIM solution is calculated using 28 EIM points and the remaining degrees of freedom are interpolated. }%
\label{fig: adv_diff}%
\end{figure}

In this section we test the quality of the DNN-EIM surrogate. We compare 
the DNN-EIM solution against the POD and DNN-EIM (s) solutions on the 
reference solution corresponding to the parameter choice 
$\theta=(\ln(A),E)=(6.4,0.11)$ which is not part of the training dataset. 
Table \ref{tab:comparison-advdiff} displays the results. First, we note 
that DNN-EIM achieves the lowest absolute error and requires the least 
number of neural network parameters. Second, we can see visually in 
Figure \ref{fig: adv_diff} that the solutions agree to a high accuracy in 
$L^2(\Omega)$-norm.

\begin{table}[h!]
\centering
\resizebox{\textwidth}{!}
{
\begin{tabular}{|c|c|c|c|c|c|}
\hline
Methods &No. ResNets & Layers & Width & No. of param. & $L^2$-error \\ [1.5ex]
\hline
\hline
POD  & 1 & 3 & 28 & 2,492 & $6.8347\cdot10^{-5}$\\
\hline
DNN-EIM (s) & 1 & 3 & 28 & 2,492 & $7.2934\cdot10^{-5}$ \\
\hline
DNN-EIM & 28 & 3 & 5 & 224 (per ResNet)  & $2.1316\cdot10^{-5}$  \\
\hline

\hline
\end{tabular}}
\caption{Comparison of POD and DNN-EIM for the advection-diffusion equation. 
Both methods have a reduced space dimension of 28. DNN-EIM overall uses fewer 
parameters and much smaller number of parameters per network. Recall that 
DNN-EIM network is trained in parallel corresponding to each EIM point.}
\label{tab:comparison-advdiff}
\end{table}

\subsection{Kuramoto-Sivashinsky Equation}
\label{sec:KSE}
The Kuramoto-Sivashinsky model is given below with periodic boundary conditions: 
\begin{subequations}
\begin{align}\label{KSPDE}
    \frac{\partial u}{\partial t} &= -\frac{\partial^4 u}{\partial x^4} - \frac{\partial^2 u}{\partial x^2} - u\frac{\partial u}{\partial x}\quad\text{ in } (0,M) \times (0,T]  ,\\
    u(0,\cdot) & =u(M,\cdot)  \quad \mbox{in } (0,T], \\
    u(\cdot,0) & = u_0(\cdot) \quad \mbox{in } (0,M).\label{KSPDE3}
\end{align}
\end{subequations}
We set $M=200$ 
and use a spectral solver with Crank-Nicolson and Adams-Bashforth (CNAB2) time stepping. 
The time-step is $dt=0.01$. The domain is discretized uniformly with meshsize $dx=\frac{1}{8}$ 
resulting in 1600 spatial degrees of freedom. The reference solution $u^{ref}$ is calculated by 
setting $u_0(x) = \cos(x)+\frac{1}{10}\cos(\frac{x}{16})(1+2\sin(\frac{x}{16}))$ and $T=10$. 

\begin{figure}[!h]
\begin{center}
\includegraphics[width=.5\textwidth]{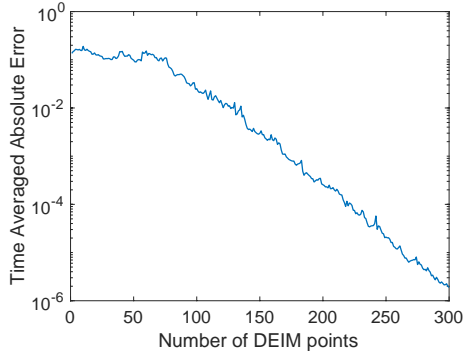}
\end{center}
\caption{Error between the EIM approximation and a reference solution
for the 1D Kuramoto-Sivashinsky equation.}%
\label{fig:ks_DEIM_err}%
\end{figure}

To generate the training samples for Step 1 of Algorithm \ref{alg:offline}, we randomly generate 
1000 initial conditions of the form $u_0^i(x) = \frac{1}{10}r_1^i + r_2^i\sin(x) + r_3^i\cos(x) 
+ r_4^i\sin(2x) + r_5^i\cos(2x)$ with $i=1,\dots,1000$ and $r^i_j$ $(j=1,\dots,5)$ is sampled 
from a standard normal distribution with mean zero and standard deviation 1. This will ensure sufficient diversity in the data. We solve 
\eqref{KSPDE} forward in time using these initial conditions and we wait 
until $200$ time units. We collect samples from the solutions every $\Delta_{NN} = 0.1$ 
time step at the EIM points 38 times. 
Since we are learning a one time step 
solution, therefore we will have 37,000 input-output pairs. The training 
data will be input output pairs of the form, 
\begin{equation}\label{eq:trainingdata}
     \bigg\{\bigg\{\bigg(\mathbf{P}^Tu_i(t_{b} + k\Delta_{NN}),\mathbf{P}^Tu_i(t_{b} + (k+1)\Delta_{NN}\bigg)\bigg\}_{k=0}^{N_s}\bigg\}_{i=1}^{N_r}.
\end{equation}
Here, $t_{b}=200$ (waiting period), $N_s=38$ is the number of snapshots per solution and $N_r=1000$ 
is the number of solutions. Next, we proceed with Steps 2-4 of Algorithm \ref{alg:offline} 
and compute the difference between EIM approximation and the reference solution. This is 
shown in Figure \ref{fig:ks_DEIM_err}. The time integration in \eqref{TARE} is done using 
10 time units. We observe the error to be approximately $10^{-6}$ when using 300 EIM 
interpolation indices or EIM points. This informed our choice to use 300 
EIM points for this particular problem.

 The ResNets in the neural network surrogate have 8 layers each with a width of 3 neurons.
\subsubsection{Neural Network Architecture}
\label{sec:nnarchKS}
In this section, we will cover the structure of our ResNet system. For each EIM interpolation 
index $p_i$ generated by Step 3 of Algorithm \ref{alg:offline}, we train one ResNet. The input to 
the ResNet is the value of our proposed algorithm (solution at the previous time-step) at $p_i$ in 
addition to the value at neighboring interpolation indices. In one dimension, assume the EIM 
interpolation indices are ordered according to their position on the real line. Then the input 
for the ResNet corresponding to the EIM interpolation index $p_i$ is $\{p_{i-1},p_i,p_{i+1}\}$, 
see Figure \ref{fig: EIM_nbd}. The output is the value at $p_i$ advanced in time with time step 
$\Delta_{NN}$.

\subsubsection{Surrogate Quality}

\begin{multicols}{2}
In this section we test the quality of the DNN-EIM surrogate.
We propagated forward the initial condition of the reference solution for one time unit  
or 10 time steps. This is equivalent to 10 neural network evaluations. We then computed 
the time average relative error between DNN-EIM and the exact solutions over one time unit
 and obtained a good approximation with absolute error $1.2 \times 10^{-2}$ as shown in Figure \ref{DNN-EIM error}. 
 We repeated the same procedure with a single ResNet 
with input and output in $\mathbb{R}^{300}$. This single ResNet has 4 layers with width 300. We observed a time averaged absolute error of $10^{6}$.

 \columnbreak

\begin{figure}[!h]
\begin{center}
\small
\begin{tikzpicture}
\node [terminator, fill=violet!30] at (0,0) (Train Data) {
Original  spatial dimension of Data $\{(\xi^i,y^i)\}_{i=1}^{N_s}$ : \textbf{1600}};
\node [terminator, fill=violet!10] at (0,-2) (Reduced) {
Reduced dimension  of Data $\{(P^T\xi^i,P^T y^i)\}_{i=1}^{N_s}$: \textbf{300} (with absolute error $10^{-6}$)};
\path [connector] (Train Data) -- (Reduced);
 \draw [arrow] (Train Data) -- node[anchor=west]  {EIM} (Reduced);
 \node [process3, fill=yellow!20] at (-3,-5) (Parameters ResNet) {\textcolor{blue}{Train on all EIM points in one ResNet} \\
 No. of Inputs: 300
 No. of Layers: 4\\
 Neuron Width: 300\\
 \textbf{Parameters} $451,200$ \\
 \textbf{Training time} $\approx 4$ Days\\
 \textbf{Surrogate error} $\approx 10^6$};
 \node [process3, fill=yellow!20] at (3,-5.0) (Reduced Parameters ResNet) {\textcolor{blue}{Train using DNN-EIM}\\
 No. of Inputs: 3
  No. of Layers: 8\\
 Neuron Width: 3\\
 \textbf{Parameters} 99 (per ResNet) and $29,700$ (total)\\
 \textbf{Training time} $\approx 1$ Hour on 34 cores.\\
 \textbf{Surrogate error} $\approx 10^{-2}$};
 \draw [arrow] (Reduced) --  (Parameters ResNet);
 \draw [arrow] (Reduced) -- (Reduced Parameters ResNet);
 \end{tikzpicture}
\end{center}

\caption{Spatial dimension reduction from EIM for the 1D Kuramoto-Sivashinsky equations. Lower Left:  Number of parameters, training time, and surrogate error for a single ResNet surrogate. Lower Right:  Number of parameters, training time, and surrogate error for DNN-EIM surrogate.}
\label{fig: Spatial parameters TIkz}
\end{figure}
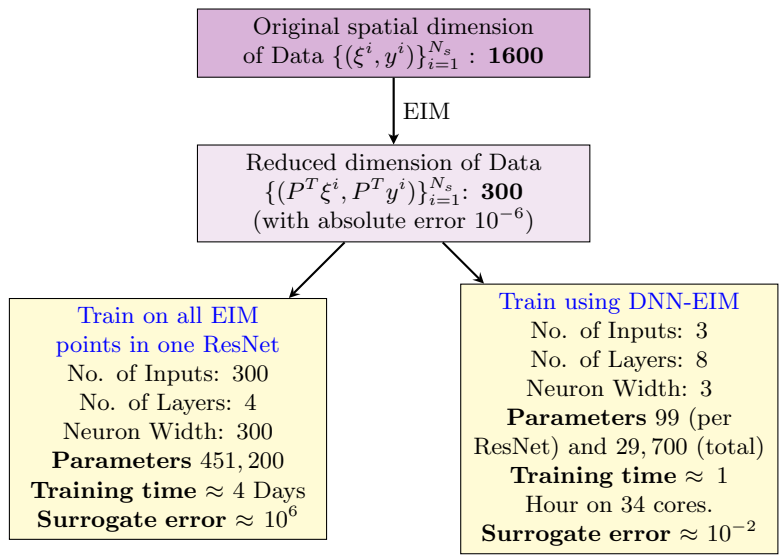

{ \centering
 \includegraphics[width=\columnwidth]{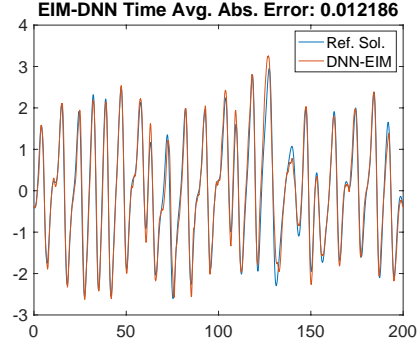}\\
 \captionof{figure}{DNN-EIM surrogate error for 1D Kuramoto-Sivashinsky equation after 10 neural network evaluations.}\label{DNN-EIM error}
}
\end{multicols}
This ResNet took approximately  4 days to train while DNN-EIM took one hour 
using 34 cores. Finally, the DNN-EIM surrogate requires training significantly 
less parameters as shown in Figure \ref{fig: Spatial parameters TIkz}. 

\section{Conclusions}
\label{sec:conclusions}
We introduced an empirical interpolation method (EIM) based approach to training DNNs called DNN-EIM. We demonstrated our approach with examples in classification, parameterized and time dependent PDEs. DNN-EIM can be trained in parallel leading to significantly smaller number of parameters to learn, reduced training times, without sacrificing the accuracy. The approach has been applied to standard MNIST dataset, parameterized advection diffusion equation, and space-time Kuramoto-Sivashinsky equation.
Future work on this topic includes how to handle more complex cases such as the unsteady 
Navier--Stokes equations. It will also be interesting to develop a convergence analysis 
of the proposed algorithm.


\bibliographystyle{siamplain}
\bibliography{references}
\end{document}

%% file: ex_shared.tex
\usepackage{lipsum}
\usepackage{amsfonts}
\usepackage{graphicx}
\usepackage{epstopdf}
\usepackage{algorithmic}
\usepackage{enumerate}
\ifpdf
  \DeclareGraphicsExtensions{.eps,.pdf,.png,.jpg}
\else
  \DeclareGraphicsExtensions{.eps}
\fi

\newsiamremark{remark}{Remark}
\newsiamremark{hypothesis}{Hypothesis}
\crefname{hypothesis}{Hypothesis}{Hypotheses}
\newsiamthm{claim}{Claim}

\headers{A Note on Dimensionality Reduction in DNNs using EIM
}{H. Antil, M. Gupta and R. Price}

\title{A Note on Dimensionality Reduction in Deep Neural Networks using Empirical Interpolation Method
\author{Harbir Antil\thanks{Department of Mathematical Sciences and The Center for Mathematics and Artificial Intelligence, George Mason University, Fairfax, VA 22030 
  (\email{hantil@gmu.edu}).}
\and Madhu Gupta\thanks{The Center for Mathematics and Artificial Intelligence, George Mason University, Fairfax, VA 22030 
  (\email{mgupta25@gmu.edu}).}
\and Randy Price\thanks{The Center for Mathematics and Artificial Intelligence, George Mason University, Fairfax, VA 22030 
  (\email{rprice25@gmu.edu}).}
}
\thanks{This work is partially supported by NSF grants DMS-2110263, DMS-1913004, the Air Force Office of Scientific Research under Award NO: FA9550-19-1-0036 and FA9550-22-1-0248.}}

\author{Dianne Doe\thanks{Imagination Corp., Chicago, IL 
  (\email{ddoe@imag.com}, \url{http://www.imag.com/\string~ddoe/}).}
\and Paul T. Frank\thanks{Department of Applied Mathematics, Fictional University, Boise, ID 
  (\email{ptfrank@fictional.edu}, \email{jesmith@fictional.edu}).}
\and Jane E. Smith\footnotemark[3]}

\usepackage{amsopn}

\makeatletter
\newcommand*{\addFileDependency}[1]{
  \typeout{(#1)}
  \@addtofilelist{#1}
  \IfFileExists{#1}{}{\typeout{No file #1.}}
}
\makeatother


%% file: paper4_EIM.bbl
\begin{thebibliography}{10}

\bibitem{antil2021deep}
{\sc H.~Antil, T.~S. Brown, R.~L{\"o}hner, F.~Togashi, and D.~Verma}, {\em Deep
  neural nets with fixed bias configuration}, arXiv preprint arXiv:2107.01308,
  (2021).

\bibitem{HAntil_DChen_SField_2018a}
{\sc H.~Antil, D.~Chen, and S.~Field}, {\em A note on qr-based model reduction:
  Algorithm, software, and gravitational wave applications}, Computing in
  Science \& Engineering, 20 (2018).

\bibitem{HAntil_HCElman_AOnwunta_DVerma_2021a}
{\sc H.~Antil, H.~C. Elman, A.~Onwunta, and D.~Verma}, {\em Novel deep neural
  networks for solving bayesian statistical inverse problems}, arXiv preprint
  arXiv:2102.03974,  (2021).

\bibitem{HAntil_SField_RHNochetto_MTiglio_2013}
{\sc H.~Antil, S.~Field, F.~Herrmann, R.~Nochetto, and M.~Tiglio}, {\em
  Two-step greedy algorithm for reduced order quadratures}, Journal of
  Scientific Computing, 57 (2013), pp.~604--637,
  \url{https://doi.org/10.1007/s10915-013-9722-z},
  \url{http://dx.doi.org/10.1007/s10915-013-9722-z}.

\bibitem{antil2014application}
{\sc H.~Antil, M.~Heinkenschloss, and D.~C. Sorensen}, {\em Application of the
  discrete empirical interpolation method to reduced order modeling of
  nonlinear and parametric systems}, Reduced order methods for modeling and
  computational reduction,  (2014), pp.~101--136.

\bibitem{MBarrault_YMaday_NCNguyen_ATPatera2004a}
{\sc M.~Barrault, Y.~Maday, N.~C. Nguyen, and A.~T. Patera}, {\em An `empirical
  interpolation' method: application to efficient reduced-basis discretization
  of partial differential equations}, C. R. Math. Acad. Sci. Paris, 339 (2004),
  pp.~667--672, \url{https://doi.org/10.1016/j.crma.2004.08.006},
  \url{http://dx.doi.org.ezproxy.rice.edu/10.1016/j.crma.2004.08.006}.

\bibitem{becker2004numerical}
{\sc R.~Becker, M.~Braack, and B.~Vexler}, {\em Numerical parameter estimation
  for chemical models in multidimensional reactive flows}, Combustion Theory
  and Modelling, 8 (2004), p.~661.

\bibitem{brown2021novel}
{\sc T.~S. Brown, H.~Antil, R.~L{\"o}hner, F.~Togashi, and D.~Verma}, {\em
  Novel dnns for stiff odes with applications to chemically reacting flows}, in
  International Conference on High Performance Computing, Springer, 2021,
  pp.~23--39.

\bibitem{chaturantabut2010nonlinear}
{\sc S.~Chaturantabut and D.~C. Sorensen}, {\em Nonlinear model reduction via
  discrete empirical interpolation}, SIAM Journal on Scientific Computing, 32
  (2010), pp.~2737--2764.

\bibitem{box}
{\sc E.~C. Cyr, M.~A. Gulian, R.~G. Patel, M.~Perego, and N.~A. Trask}, {\em
  Robust training and initialization of deep neural networks: An adaptive basis
  viewpoint}, in Mathematical and Scientific Machine Learning, PMLR, 2020,
  pp.~512--536.

\bibitem{franco2023deep}
{\sc N.~Franco, A.~Manzoni, and P.~Zunino}, {\em A deep learning approach to
  reduced order modelling of parameter dependent partial differential
  equations}, Mathematics of Computation, 92 (2023), pp.~483--524.

\bibitem{galbally2010non}
{\sc D.~Galbally, K.~Fidkowski, K.~Willcox, and O.~Ghattas}, {\em Non-linear
  model reduction for uncertainty quantification in large-scale inverse
  problems}, International journal for numerical methods in engineering, 81
  (2010), pp.~1581--1608.

\bibitem{he2016deep}
{\sc K.~He, X.~Zhang, S.~Ren, and J.~Sun}, {\em Deep residual learning for
  image recognition}, in Proceedings of the IEEE conference on computer vision
  and pattern recognition, 2016, pp.~770--778.

\bibitem{JSHesthaven_GRozza_BStamm_2016a}
{\sc J.~S. Hesthaven, G.~Rozza, and B.~Stamm}, {\em Certified reduced basis
  methods for parametrized partial differential equations}, SpringerBriefs in
  Mathematics, Springer, Cham; BCAM Basque Center for Applied Mathematics,
  Bilbao, 2016, \url{https://doi.org/10.1007/978-3-319-22470-1},
  \url{https://doi.org/10.1007/978-3-319-22470-1}.
\newblock BCAM SpringerBriefs.

\bibitem{hesthaven2018non}
{\sc J.~S. Hesthaven and S.~Ubbiali}, {\em Non-intrusive reduced order modeling
  of nonlinear problems using neural networks}, Journal of Computational
  Physics, 363 (2018), pp.~55--78.

\bibitem{YMaday_2009a}
{\sc Y.~Maday, N.~C. Nguyen, A.~T. Patera, and G.~S.~H. Pau}, {\em A general
  multipurpose interpolation procedure: the magic points}, Commun. Pure Appl.
  Anal., 8 (2009), pp.~383--404, \url{https://doi.org/10.3934/cpaa.2009.8.383},
  \url{http://dx.doi.org/10.3934/cpaa.2009.8.383}.

\bibitem{maulik2021reduced}
{\sc R.~Maulik, B.~Lusch, and P.~Balaprakash}, {\em Reduced-order modeling of
  advection-dominated systems with recurrent neural networks and convolutional
  autoencoders}, Physics of Fluids, 33 (2021), p.~037106.

\bibitem{LRuthotto_EHaber_2019a}
{\sc L.~Ruthotto and E.~Haber}, {\em Deep neural networks motivated by partial
  differential equations}, J. Math. Imaging Vision,  (2019),
  \url{https://doi.org/10.1007/s10851-019-00903-1}.

\bibitem{salvador2021non}
{\sc M.~Salvador, L.~Dede, and A.~Manzoni}, {\em Non intrusive reduced order
  modeling of parametrized pdes by kernel pod and neural networks}, Computers
  \& Mathematics with Applications, 104 (2021), pp.~1--13.

\bibitem{san2019artificial}
{\sc O.~San, R.~Maulik, and M.~Ahmed}, {\em An artificial neural network
  framework for reduced order modeling of transient flows}, Communications in
  Nonlinear Science and Numerical Simulation, 77 (2019), pp.~271--287.

\bibitem{MR3732946}
{\sc J.~A. Tropp, A.~Yurtsever, M.~Udell, and V.~Cevher}, {\em Practical
  sketching algorithms for low-rank matrix approximation}, SIAM J. Matrix Anal.
  Appl., 38 (2017), pp.~1454--1485, \url{https://doi.org/10.1137/17M1111590},
  \url{https://doi.org/10.1137/17M1111590}.

\end{thebibliography}
